\documentclass[conference]{IEEEtran}
\IEEEoverridecommandlockouts

\usepackage{cite}
\usepackage{amsmath,amssymb,amsfonts}
\usepackage{algorithmic}
\usepackage{graphicx}
\usepackage{textcomp}
\usepackage{xcolor}
\usepackage{caption}
\usepackage{subcaption}

\def\BibTeX{{\rm B\kern-.05em{\sc i\kern-.025em b}\kern-.08em
    T\kern-.1667em\lower.7ex\hbox{E}\kern-.125emX}}

\usepackage{fancyhdr}
\usepackage{etoolbox}

\fancypagestyle{firstpage}{
    \fancyhf{}
    \fancyhead[L]{\small
    This is the author’s accepted manuscript of a paper accepted at the IEEE World AI IoT Congress 2026. The final published version will be available via IEEE Xplore.
    }

}
    
\begin{document}

\title{Making Brain-Computer Interfaces More Secure\\

}

\author{
\IEEEauthorblockN{Md Fahimul Kabir Chowdhury\textsuperscript{1} and Gahangir Hossain\textsuperscript{2}}
\IEEEauthorblockA{
\textsuperscript{1}\textit{Department of Computer Science and Engineering, University of North Texas, USA} \\
\textsuperscript{2}\textit{Department of Data Science, University of North Texas, USA} \\
MdFahimulKabirChowdhury@my.unt.edu, Gahangir.Hossain@unt.edu
}
}

\maketitle
\thispagestyle{firstpage}

\begin{abstract}
The development of brain–computer interfaces (BCIs) based on electroencephalograms (EEGs) has advanced significantly mainly to machine learning. Although the majority of earlier research has been on increasing classification accuracy, relatively little focus has been placed on security and robustness. According to recent research, EEG-based BCIs are susceptible to adversarial attacks, which can cause misdiagnosis due to minute, well-crafted disturbances. Evaluating model robustness against such perturbations is therefore critical for ensuring reliable deployment. In this study, we propose a lightweight custom Convolutional Neural Network (CNN) architecture to investigate adversarial robustness in EEG-based BCIs. The suggested method is assessed using two EEG datasets and contrasted with three novel CNN models tailored to EEG, namely EEGNet, DeepConvNet, and SleepEEGNet, under gradient-based adversarial attack scenarios. According to experimental findings, the suggested model continuously performs better in classification under adversarial perturbations compared to baseline models, indicating improved robustness. These findings highlight the potential of lightweight architectures for enhancing the reliability of EEG-based BCI systems under adversarial conditions.
\end{abstract}

\begin{IEEEkeywords}
Adversarial training, Brain-computer interface, electroencephalogram, adversarial attack, FGSM
\end{IEEEkeywords}

\section{Introduction}
\begin{figure}[htbp]
\centering
\includegraphics[width=\linewidth]{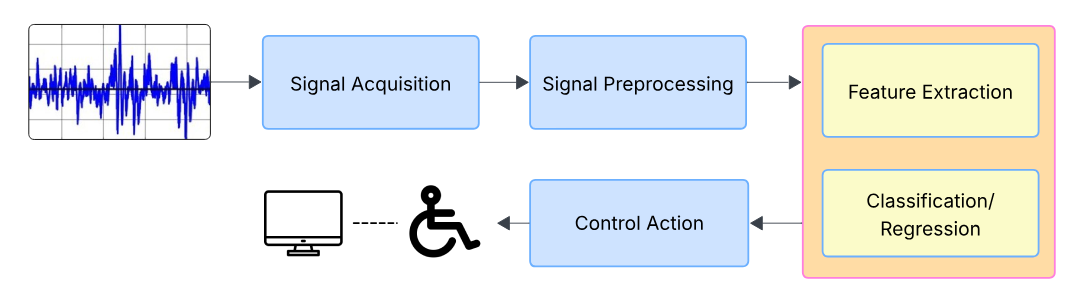}
\caption{Overview of a EEG-based BCIs system.}
\label{fig:framework_overview}
\end{figure}

The human brain and external equipment may communicate directly with the help of a brain-computer interface (BCI) and has been widely explored in fields such as neuroscience, neural engineering, and clinical recovery~\cite{ienca2018brain}. Because it is inexpensive, noninvasive, and simple to use, electroencephalography (EEG) has become the most popular brain recording method for BCIs ~\cite{varbu2022past}. The four primary parts of a typical EEG-based BCI architecture are signal acquisition, signal preprocessing, machine learning, and control execution, as shown in Fig.~\ref{fig:framework_overview}. Conventional methods at the machine learning stage usually entail manual feature extraction, classification, or regression.

In applications such as human behavior analysis, brain signal analysis has been widely used~\cite{essahraui2025human,hossain2018spatial}, neuro-assistive systems including seizure prediction~\cite{myers2016seizure,myers2022dual,ghane2020learning}, and more recently, cybersecurity applications~\cite{hossaincogntive,hossain2016pattern}. In EEG classification, signal decomposition techniques are among the most frequently applied approaches~\cite{sadiq2019motor,Korani2025cnn}. For example, Sadiq \textit{et al.}~\cite{han2017electrocardiogram} introduced a multivariate variational mode decomposition (MVMD) method, while Yu \textit{et al.}~\cite{yu2021new}, for EEG-based classification tasks, empirical Fourier decomposition (EFD) and its updated form (IEFD) were used. Signal denoising has also been widely recognized as a critical step for enhancing classification accuracy~\cite{sadiq2022alcoholic}. In this context, Sadiq \textit{et al.}~\cite{sadiq2020motor} proposed the multiscale principal component analysis (MSPCA) technique which can reduce noise in EEG recordings. Graphical feature-based methods have become a viable avenue for revealing hidden patterns in EEG data in more recent times.~\cite{Korani2025cnn,akbari2023recognizing}. 

Beyond these traditional techniques, deep learning models have demonstrated strong performance by learning hierarchical representations of features automatically, doing away with the requirement for manually created feature extraction\cite{chowdhury2022development,fahimul2026cnn}. However, while prior research has primarily focused on improving classification accuracy, relatively security and resilience have received little attention in EEG-based BCIs. Existing studies often overlook how adversarial perturbations can compromise model reliability, making adversarial resilience an underexplored yet critical area of research.

\section{Related Work}

\subsection{Attack Approaches}
Adversarial attacks are categorized according to the level of exposure the attacker has to the target architecture, where black-box, gray-box, and white-box attacks are the three types. Because they presume total knowledge of the model's structure and variables, white-box attacks are thought to be the most effective. Consequently, evaluating model robustness under white-box settings provides a strong benchmark for worst-case vulnerability. In this study, we concentrate on two popular white-box attack methods: FGSM and PGD. Adversarial examples are purposefully distorted inputs intended to deceive machine learning models while remaining nearly indistinguishable from the original data.

\subsubsection{Fast Gradient Sign Method (FGSM)}
An attack method called FGSM uses gradient information to create adversarial instances~\cite{goodfellow2014explaining}. It creates adversarial samples by optimizing the model's loss function in a single step. Specifically, a small perturbation $\epsilon$ is applied in the direction of the gradient, producing an adversarial example as follows:

\begin{equation}
x_{adv} = x + \epsilon \cdot \text{sign}(\nabla_x J(\theta, x, y)),
\end{equation}

where $J(\theta, x, y)$ is the loss function, $x$ is the initial input, and $\theta$ stands for the model parameters. The $\text{sign}(\cdot)$ function ensures that the perturbation follows the gradient direction.

For EEG-based BCI regression, Meng \textit{et al.}~\cite{meng2019white} presented one of the first research on adversarial attacks, introducing white-box targeted attack strategies that effectively alter regression outputs. Zhang \textit{et al.} ~\cite{zhang2019vulnerability} proposed an unsupervised variant, termed Unsupervised FGSM (UFGSM), which replaces true labels with predicted labels for EEG-based BCIs. Their results demonstrate that such attacks are both effective and transferable across models, highlighting critical security concerns.

To jointly perform classification and adversarial discrimination, Aissa \textit{et al.}~\cite{aissa2023robust} suggested using adversarial training for a hierarchical neural network. With BCI Competition IV-2a dataset, the model was evaluated under FGSM attacks, the method achieved 99.92\% accuracy and a 0.9985 Cohen’s Kappa score, demonstrating strong robustness.

\subsubsection{Projected Gradient Descent (PGD)}
PGD is an iterative extension of FGSM~\cite{madry2017towards}, which performs multiple small update steps while constraining the perturbation within a predefined range. It begins with a random initialization near the original input and then iteratively refines the adversarial example:

\begin{equation}
x_{adv}^0 = x + \xi,
\end{equation}

\begin{equation}
x_{adv}^i = \text{Proj}_{x,\epsilon} \big(x_{adv}^{i-1} + \alpha \cdot \text{sign}(\nabla_{x_{adv}^{i-1}} J(\theta, x_{adv}^{i-1}, y))\big),
\end{equation}

where random noise is denotes by $\xi \in U(-\epsilon, \epsilon)$, $\alpha$ is the step size, $i = 1,2,\ldots,n_{iter}$, and $\text{Proj}_{x,\epsilon}$ projects the adversarial sample onto the $\epsilon$-bounded neighborhood of the original input under the $l_\infty$ norm.

Feng \textit{et al.}~\cite{feng2021saga} introduced SAGA, a framework for EEG analytics that perturbs only a small subset of channels and time steps using an adaptive mask combined with a PGD-based solver. Which is also called as a sparse adversarial attack. Experimental results show that SAGA can cause an average accuracy drop of 77.02\% by modifying only 5\% of the data, highlighting the significant vulnerability of EEG-based BCI systems.






\section{Methodology}

\subsection{Basic Idea}
\begin{figure*}[t]
    \centering
    \includegraphics[width=\textwidth]{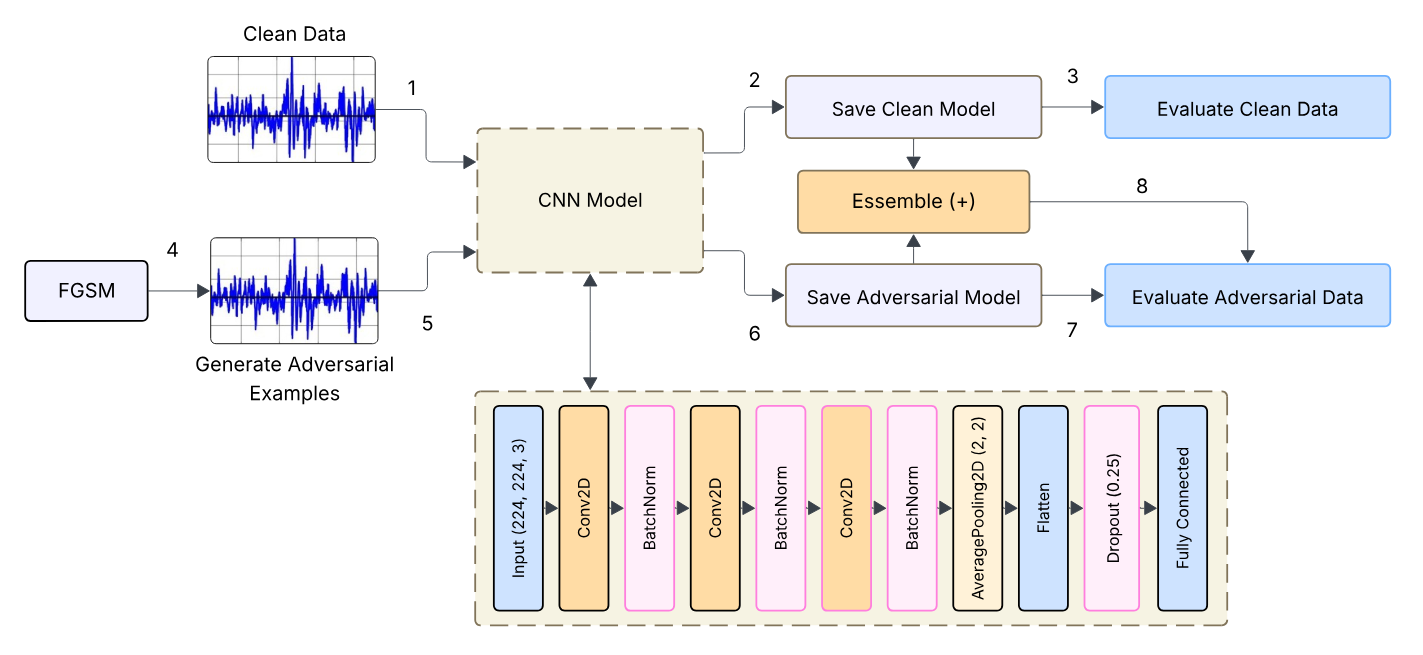}
    \caption{The process of preprocessing EEG signals and generating time-frequency representations to develop the proposed lightweight CNN architecture.}
    \label{blockD}
\end{figure*}

The overall workflow of this study follows a structured pipeline consisting of data preprocessing, fold-wise model training, ensemble integration, and robustness evaluation. We consider white-box, untargeted adversarial attack scenarios, where the model parameters are fully accessible to the attacker and aims to induce misclassification without specifying a target class. FGSM and PGD are employed under this setting to evaluate worst-case robustness. Figure~\ref{blockD} presents a high-level depiction of the proposed framework.

Every EEG spectrogram picture is normalized to the range [0,1] and scaled to a set resolution of 224 X 224 pixels. To ensure reliable performance estimation, we use stratified $K$-fold cross-validation ($K=10$), where dataset $\mathcal{D}$ is partitioned into $K$ mutually exclusive subsets $\{\mathcal{D}_1, \mathcal{D}_2, \dots, \mathcal{D}_K\}$. For each fold $k$, a model $f_k(\cdot;\theta_k)$ is trained on $\mathcal{D} \setminus \mathcal{D}_k$ and evaluated on $\mathcal{D}_k$.

The categorical cross-entropy loss is optimized by the training procedure:
\begin{equation}
\mathcal{L} = - \frac{1}{N} \sum_{i=1}^{N} \sum_{c=1}^{C} y_{i,c} \log \hat{y}_{i,c},
\end{equation}
where $\hat{y}_{i,c}$ is the predicted probability, $y_{i,c}$ is the ground-truth label for class $c$, $N$ is the number of sample sizes, and $C=4$ is the number of categories. The Adam optimizer is used for optimization, with an initial learning rate of $1\times10^{-4}$, early halting, and learning rate scheduling to avoid overfitting.

Upon completion of fold-wise training, the $K$ models are integrated into an ensemble framework. Let $\hat{y}^{(k)}$ denote the class-probability output of model $f_k$. The final ensemble prediction is obtained via average fusion:
\begin{equation}
\hat{y} = \frac{1}{K} \sum_{k=1}^{K} \hat{y}^{(k)}.
\end{equation}
The ensemble is then evaluated as a unified model.

To assess robustness, both individual models and the ensemble are evaluated on clean and adversarially perturbed test sets. Accuracy, precision, recall, F1-score, confusion matrices, and ROC-AUC are used to evaluate performance. Results are reported per fold and averaged across folds, with standard deviation included to quantify stability. Training histories (loss and accuracy curves) are also analyzed to ensure consistent convergence across all folds.

\subsection{Proposed CNN Architecture}
Throughout all experiments, we employ a lightweight Convolutional Neural Network (CNN) designed to balance classification performance and computational efficiency. The network takes spectrogram images of size $224 \times 224 \times 3$ as input and processes them through a sequence of convolutional and pooling operations.

The first two convolutional blocks use $3\times3$ filters with 8 and 16 channels, respectively, to improve feature extraction and stabilize training, ReLU activation and batch normalization are then applied. A max-pooling layer reduces spatial resolution, enabling the model to capture higher-level patterns while controlling parameter growth. Followed by a convolutional layer with 32 filters, accompanied by batch normalization, extracts more discriminative features, after which average pooling further compresses the learned representations.

To reduce overfitting, the resultant feature maps are flattened into a one-dimensional vector and regularized with a dropout layer at a rate of 0.25. In the final stage, a fully connected dense layer consisting of four output neurons applies a softmax activation function to generate class probabilities.

Our proposed lightweight CNN is well-suited for adversarial settings due to its reduced parameter complexity and controlled feature extraction process, which can help limit overfitting to high-frequency or noise-like patterns often exploited by adversarial perturbations. Compared to deeper architectures such as DeepConvNet and specialized models like EEGNet and SleepEEGNet, the proposed model offers a balanced trade-off between generalization, computational efficiency, and robustness under adversarial conditions. A summary of the architecture is illustrated in Fig.~\ref{blockD}.

\section{Experimental Setup}

\subsection{Datasets}
Two EEG datasets are used in this study: a private dataset and a benchmark dataset that is accessible to the public. This combination enables evaluation across both standard motor imagery tasks (MI4) and clinically relevant EEG signals (rTMS dataset). As a result, the proposed approach is assessed under both controlled experimental conditions and real-world medical scenarios, enhancing the practical relevance and generalizability of the findings. A description of these datasets and their preparation methods is given below.

\subsubsection{Four-Class Motor Imagery Dataset (MI4)}
The MI4 dataset~\cite{tangermann2012review}, also known as Dataset 2a from BCI Competition IV, where the data was gathered from nine participants over the course of two sessions on separate days. The left hand, right hand, foot, and tongue are its four motor imagery labels. 22 channels of EEG waves were captured at a 250 Hz sampling rate. For analysis, tests were band-pass filtered between 8 and 32 Hz after data from 0 to 4 seconds following each imagining trigger were removed. For each lesson, each participant provided 144 EEG epochs.

\subsubsection{rTMS Therapy EEG Dataset}
The rTMS dataset was collected from 15 individuals at Atieh Hospital, Tehran, Iran, diagnosed with depression and undergoing rTMS treatment prescribed by clinical specialists. Treatment outcomes were evaluated using Beck Depression Inventory (BDI) scores measured before and four weeks after therapy. Patients were labeled as responders (R) if their BDI scores decreased by at least 50\%, and non-responders (NR) otherwise, with classifications verified by clinicians.

EEG recordings were obtained using the standard 10-20 electrode system with 19 scalp channels, sampled at 500 Hz. Signals were segmented into fixed-length windows of 1024 samples (approximately 2.05 seconds) for consistency. The A1-A2 reference channel pair was excluded from this dataset. Although the dataset size is relatively limited, to improve the reliability and stability of the results, stratified cross-validation and repeated experiments are employed.

\subsection{CNN Models for EEG Classification}
Along side with our proposed lightweight custom CNN. In this experiment, the following three convolutional neural network models are used, which are common in EEG signals:

\subsubsection{EEGNet} 
A small convolutional neural network called EEGNet \cite{lawhern2018eegnet} was created especially for EEG-based brain–computer interactions. It is appropriate for limited data sets and real-time applications because it effectively extracts both temporal and spatial characteristics from EEG signals using depthwise and separable convolutions.

\subsubsection{DeepConvNet} 
DeepConvNet \cite{ding2017energy} is a deeper CNN architecture for EEG decoding. It combines multiple convolutional and pooling layers to hierarchically acquire spatial and temporal representations from unprocessed EEG data. Its deeper structure enables it to capture complex patterns, but it typically requires more data and computational resources.

\subsubsection{SleepEEGNet} 
SleepEEGNet \cite{mousavi2019sleepeegnet} is a lightweight CNN tailored for automatic sleep stage classification. The network employs multiple temporal convolutional layers followed by spatial filtering to maintain a modest model size for practical application while extracting biased features from multi-channel EEG data.

\subsection{Experimental Setup}

A within-subject experimental design is adopted, where 10\% of each participant's EEG data is used for testing and 90\% is used for training. Since smaller dataset tends to overfit easily, an early halting approach with a patience of six epochs is used to avoid overfitting. Model performance is evaluated using both average classification accuracy and balanced accuracy across all subjects. The CNN models are trained using the categorical cross-entropy loss function in combination with the Adam optimizer. To ensure reproducibility, every experiment is conducted 10 times, and the average outcomes are presented.

For adversarial evaluation, perturbation strengths ($\epsilon$) of 0.1, 0.3, and 0.5 are applied across all selected attack methods. FGSM attacks are performed under the $l_\infty$-norm constraint. The baseline experiment involves classification on clean datasets using four models under identical training settings to ensure fair comparison. With a batch size of 32 and a starting learning rate of $1\times10^{-4}$, training is carried out for 100 epochs, with dynamic learning rate scheduling applied. All implementations are carried out using TensorFlow~2.14 with Python~3.10.

\section{Results}

Table~\ref{tab:fgsm_results} presents the classification performance of four CNN models under FGSM adversarial attacks. The baseline accuracies indicate that the proposed CNN achieves higher performance (88.21\%) compared to EEGNet (42.35\%), DeepConvNet (47.66\%), and SleepEEGNet (24.91\%). After applying FGSM perturbations, accuracy drops significantly across all models; however, the proposed CNN maintains a substantially higher accuracy (73.02\%) relative to the other networks, which fall below 7\%.

\begin{table*}[htbp]
\centering
\caption{Performance of Models under FGSM Attack}
\label{tab:fgsm_results}
\begin{tabular}{|c|c|c|c|c|c|c|c|}
\hline
\textbf{Models} & \textbf{Baseline Acc} & \textbf{After Attack Acc} & \textbf{$\epsilon$ = 0.1} & \textbf{$\epsilon$ = 0.3} & \textbf{$\epsilon$ = 0.5} & \textbf{Avg Kappa} & \textbf{Avg Accuracy} \\ \hline
SleepEEGNet & 24.91 & 1.36 & 44.66 & 25.95 & 24.02 & 32.09 & 31.54 \\ \hline
EEGNet      & 42.35 & 5.27 & 50.76 & 34.43 & 31.38 & 36.66 & 38.85 \\ \hline
DeepConvNet & 47.66 & 6.17 & 99.44 & 95.18 & 90.51 & 93.39 & 95.04 \\ \hline
\textbf{Proposed CNN} & \textbf{88.21} & \textbf{73.02} & \textbf{100.00} & \textbf{99.99} & \textbf{99.94} & \textbf{99.96} & \textbf{99.97} \\ \hline
\end{tabular}
\end{table*}

When varying perturbation strengths ($\epsilon = 0.1, 0.3, 0.5$), DeepConvNet and the proposed CNN demonstrate consistently high robustness, maintaining accuracies above 90\% across all cases. In contrast, EEGNet and SleepEEGNet show moderate performance at lower perturbation levels, reaching 50.76\% and 44.66\%, respectively, but degrade more noticeably as $\epsilon$ increases. The average Cohen’s Kappa further supports this trend, with the proposed CNN (99.96) and DeepConvNet (93.39) outperforming EEGNet (36.66) and SleepEEGNet (32.09).

\begin{figure}[htbp]
\centering
\includegraphics[width=\linewidth]{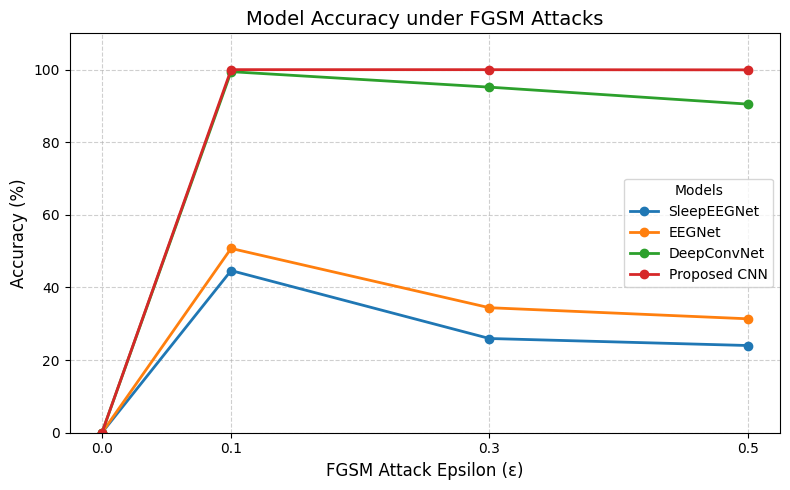}
\caption{Accuracy of different CNN models under FGSM attacks with varying perturbation strengths ($\epsilon$).}
\label{fig:acc_comparison}
\end{figure}

\begin{figure*}[htbp]
    \centering
    \subfloat[Before FGSM attack]{%
        \includegraphics[width=0.3\linewidth]{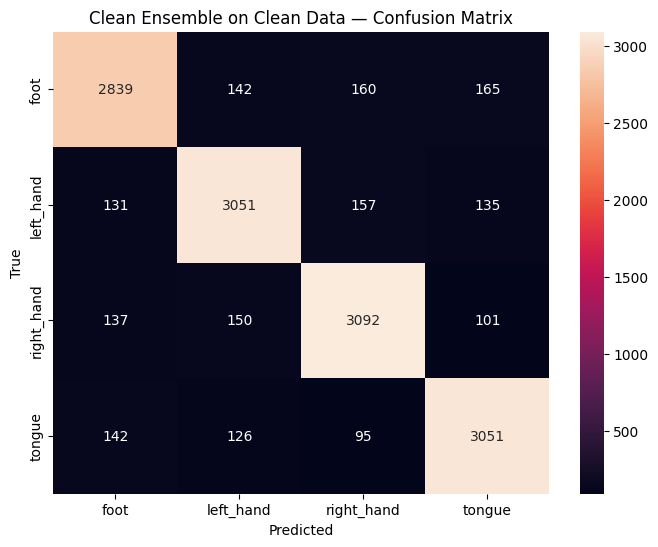}
    }
    \hfil
    \subfloat[After FGSM attack]{%
        \includegraphics[width=0.3\linewidth]{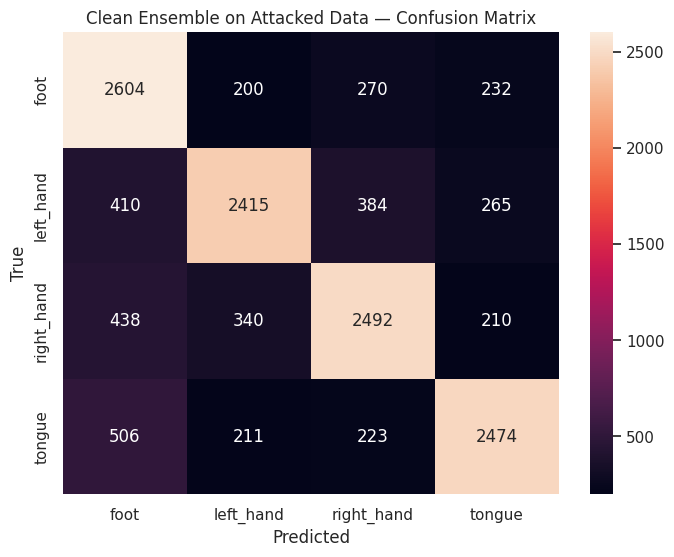}
    }
    \hfil
    \subfloat[After FGSM attack with Robust Model]{%
        \includegraphics[width=0.3\linewidth]{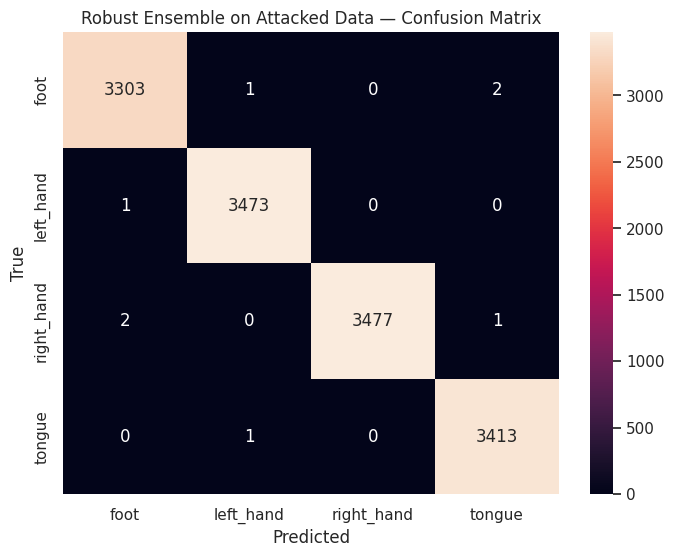}
    }
    \caption{Confusion matrices of the proposed CNN before and after applying FGSM attack.}
    \label{Fig:fgsm_cm}
\end{figure*}

Figure~\ref{fig:acc_comparison} illustrates the impact of FGSM adversarial perturbations at varying strengths on model accuracy, showing that the proposed CNN maintains comparatively higher robustness across different attack intensities. Also figure~\ref{Fig:fgsm_cm} shows the classification differences of before and after FGSM attack through confusion matrix plot with proposed CNN.

While the proposed CNN achieves very high performance under adversarial conditions, we acknowledge that such results may appear unusually strong. This behavior can be attributed to the combination of (i) spectrogram-based representations that enhance discriminative patterns, (ii) consistent training settings across all models, and (iii) the use of ensemble averaging, which improves stability and reduces variance across folds. Importantly, all baseline models (EEGNet, DeepConvNet, SleepEEGNet) are trained under identical preprocessing, training schedules, and evaluation protocols to ensure a fair comparison. The observed performance differences therefore reflect variations in architectural behavior under adversarial perturbations rather than inconsistencies in training.

Overall, these results indicate that while FGSM attacks significantly degrade performance in standard EEG-specific networks, the proposed lightweight CNN demonstrates improved resilience under adversarial perturbations and consistently outperforms the baseline models. These results demonstrate how reliable and computationally viable designs may improve the dependability of EEG-based BCI systems in practical applications including healthcare diagnosis, assistive technologies, and neurorehabilitation.

\section{Conclusion}

In this work, we adapted adversarial evaluation techniques originally developed for computer vision to EEG-based brain-computer interfaces (BCIs). Using the suggested lightweight CNN and three cutting-edge EEG-based CNN models, we assessed model performance under gradient-based white-box attacks, such FGSM, on two EEG datasets. The results demonstrate that classification models can maintain varying levels of robustness under adversarial perturbations, with some architectures being more resilient than others.

Notably, the suggested lightweight CNN achieves consistently high performance under adversarial conditions, reaching an average accuracy of 99.97\%. This suggests that appropriately designed lightweight architectures can improve robustness while maintaining computational efficiency. However, these results should be interpreted within the context of the selected datasets and experimental setup.

Future work will focus on the following directions:
\begin{enumerate}
    \item Exploring additional robustness evaluation metrics to capture diverse characteristics of adversarial perturbations and provide deeper insights into model behavior.
    \item Designing and integrating complementary feature representations to improve model generalization under adversarial conditions.
    \item Assessing the model's resilience to a greater variety of attack techniques, such as transfer-based and black-box attacks.
    \item Investigating dedicated defense mechanisms, such as adversarial training and input preprocessing, to further enhance system reliability.
\end{enumerate}

\bibliographystyle{IEEEtran}
\bibliography{ref}

\end{document}